\documentclass{amia}
\usepackage{graphicx}
\usepackage[labelfont=bf]{caption}
\usepackage[superscript,nomove]{cite}
\usepackage{color}

\begin{document}

\title{Deep Health Care Text Classification}

\author{Vinayakumar R, Barathi Ganesh HB, Anand Kumar M, Soman KP}

\institutes{
    Center for Computational Engineering and Networking (CEN), Amrita School of Engineering Coimbatore,
Amrita Vishwa Vidyapeetham, Amrita University, India, \\
$vinayakumarr77@gmail.com$,$barathiganesh.hb@gmail.com$, $m\_anandkumar@cb.amrita.edu$, $kp\_soman@amrita.edu$\\
}

\maketitle

\noindent{\bf Abstract}

\textit{Health related social media mining is a valuable apparatus for the early recognition of the diverse antagonistic medicinal conditions. Mostly, the existing methods are based on machine learning with knowledge-based learning. This working note presents the Recurrent neural network (RNN) and Long short-term memory (LSTM) based embedding for automatic health text classification in the social media mining. For each task, two systems are built and that classify the tweet at the tweet level. RNN and LSTM are used for extracting features and non-linear activation function at the last layer facilitates to distinguish the tweets of different categories. The experiments are conducted on 2nd Social Media Mining for Health Applications Shared Task at AMIA 2017. The experiment results are considerable; however the proposed method is appropriate for the health text classification. This is primarily due to the reason that, it doesn't rely on any feature engineering mechanisms.}

\section*{Introduction}
With the expansion of micro blogging platforms such as Twitter, the Internet is progressively being utilized to spread health information instead of similarly as a wellspring of data \cite{ref1, ref2}. Twitter allows users to share their status messages typically called as tweets, restricted to 140 characters. Most of the time, these tweets expresses the opinions about the topics. Thus analysis of tweets has been considered as a significant task in many of the applications, here for health related applications.

Health text classification is taken into account a special case of text classification. The existing methods have used machine learning methods with feature engineering. Most commonly used features are n-grams, parts-of-speech tags, term frequency-inverse document frequency, semantic features such as mentions of chemical substance and disease, WordNet synsets, adverse drug reaction lexicon, etc \cite{ref3,ref4,ref5,ref6, ref15}. In \cite{ref6,ref7} proposed ensemble based approach for classifying the adverse drug reactions tweets. 

Recently, the deep learning methods have performed well \cite{ref8} and used in many tasks mainly due to that it doesn't rely on any feature engineering mechanism. However, the performance of deep learning methods implicitly relies on the large amount of raw data sets. To make use of unlabeled data, \cite{ref9} proposed semi-supervised approach based on Convolutional neural network for adverse drug event detection. Though the data sets of task 1 and task 2 are limited, this paper proposes RNN and LSTM based embedding method.

\section*{Background and hyper parameter selection}

This section discusses the concepts of tweet representation and deep learning algorithms particularly recurrent neural network (RNN) and long short-term memory (LSTM) in a mathematical way.

\subsection*{Tweet representation}
Representation of tweets typically called as tweet encoding. This contains two steps. The tweets are tokenized to words during the first step. Moreover, all words are transformed to lower-case. In second step, a dictionary is formed by assigning a unique key for each word in a tweet. The unknown words in a tweet are assigned to default key 0. To retain the word order in a tweet, each word is replaced by a unique number according to a dictionary. Each tweet vector sequence is made to same length by choosing the particular length. The tweet sequences that are too long than the particular length are discarded and too short are padded by zeros.  This type of word vector representation is passed as input to the word embedding layer. For task 1, the maximum tweet sequence length is 35. Thus the train matrix of shape 6725*35, valid matrix of shape 3535*35 is passed as input to an embedding layer. For task 2, the maximum tweet sequence length is 34. Thus the train matrix of shape 1065*34, valid matrix of shape 712*34 is passed as input to an embedding layer. Word embedding layer transforms the word vector to the word embedding by using the following mathematical operation.

\begin{equation}
   Input-shape * weights-of-word-embedding = (nb-words, word-embedding-dimension)
\end{equation}

where input-shape = (nb-words, vocabulary-size), nb-words denotes the number of top words, vocabulary-size denotes the number of unique words, weights-of-word-embedding = (vocabulary-size, word-embedding-dimension), word-embedding-dimension denotes the size of word embedding vector. This kind of mathematical operation transforms the discrete number to its vectors of continuous numbers. This word embedding layer captures the semantic meaning of the tweet sequence by mapping them in to a high dimensional geometric space. This high dimensional geometric space is called as an embedding space. If an embedding is properly learnt the semantics of the tweet by encoding as a real valued vectors, then the similar tweets appear in a same cluster with close to each other in a high dimensional geometric space. To select optimal parameter for the embedding size, two trails of experiments are run with embedding size 128, 256 and 512. For each experiment, learning rate is set to 0.01. An experiment with embedding size 512 performed well in both the RNN and LSTM networks. Thus for the rest of the experiments embedding size is set to 512. The embedding layer output vector is further passed to RNN and its variant LSTM layer. RNN and LSTM obtain the optimal feature representation and those feature representation are passed to the dropout layer. Dropout layer contains 0.1 which removes the neurons and its connections randomly. This acts as a regularization parameter. In task 1 the output layer contains   $sigmoid$ activation function and  $softmax$ activation function for task 2.

\subsection*{Recurrent neural network (RNN) and it’s variant}

Recurrent neural network (RNN) was an enhanced model of feed forward network (FFN) introduced in 1990 \cite{ref10}. The input sequences ${x_T}$ of arbitrary length are passed to RNN and a transition function $tf$ maps them into hidden state vector $h{i_{t - 1}}$ recursively. The hidden state vector $h{i_{t - 1}}$ are calculated based on the transition function $tf$ of present input sequence ${x_T}$ and previous hidden state vector $h{i_{t - 1}}$. This can be mathematically formulated as follows

\begin{equation}
    h{i_t} = \,\left\{ \begin{array}{l}
0\,\,\,\,\,\,\,\,\,\,\,\,\,\,\,\,\,\,\,\,\,\,\,\,\,t = 0\,\\
tf(h{i_{t - 1}},{x_t})\,\,\,{\rm{otherwise}}
\end{array} \right\}
\end{equation}

This kind of transition function  results in vanishing and exploding gradient issue while training \cite{ref11}. To alleviate, LSTM was introduced \cite{ref11,ref12,ref13}. LSTM network contains a special unit typically called as a memory block. A memory block composed of a memory cell $m$ and set of gating functions such as input gate $(ig)$, forget gate $(fr)$ and output gate $(og)$ to control the states of a memory cell. The transition function $tf$ for each LSTM units is defined below

\begin{equation}
    i{g_t} = \sigma ({w_{ig}}{x_t} + \,{P_{ig}}h{i_{t - 1}} + \,{Q_{ig}}{m_{t - 1}} + {b_{ig}})
\end{equation}

\begin{equation}
    f{g_t} = \sigma ({w_{fg}}{x_t} + \,{P_{fg}}h{i_{t - 1}} + \,{Q_{fg}}{m_{t - 1}} + {b_{fg}})
\end{equation}

\begin{equation}
    o{g_t} = \sigma ({w_{og}}{x_t} + \,{P_{og}}h{i_{t - 1}} + \,{Q_{og}}{m_{t - 1}} + {b_{og}})
\end{equation}

\begin{equation}
    m{1_t} = \tanh ({w_m}{x_t} + \,{P_m}h{i_{t - 1}} + {b_m})
\end{equation}

\begin{equation}
    {m_t} = \,fg_t^i\, \odot \,{m_{t - 1}} + \,i{g_t} \odot m1
\end{equation}

\begin{equation}
    h{i_t} = \,o{g_t}\, \odot \tanh ({m_t})
\end{equation}

where ${x_t}$ is the input at time step $t$, $P$ and $Q$ are weight parameters, $\sigma$ is sigmoid activation function,$\odot$ denotes element-wise multiplication.

\section*{Experiments}

This section discusses the data set details of task 1 and task 2 and followed by experiments related to parameter tuning. 

Task 1 is aims at classifying the twitter posts to either the existence of adverse drug reaction (ADR) or not. Task 2 aims at classifying the twitter posts to personal medication intake, possible medication intake or non-intake. The data sets for all two tasks are provided by shared task committee and the detailed statistics of them are reported in Table 1 and Table 2. Each task data set is composed of train, validation and test data sets.

\begin{table}[!ht]
\centering
\caption{Task 1 Data Statistics}
\label{my-label}
\begin{tabular}{|l|l|l|l|l|}
\hline
\textbf{Data} & \textbf{\begin{tabular}[c]{@{}l@{}}Total \#\\ Tweets\end{tabular}} & \textbf{\begin{tabular}[c]{@{}l@{}}Total \#\\ Classes\end{tabular}} & \textbf{\begin{tabular}[c]{@{}l@{}}\# ADR Mentioned\\ Tweets\end{tabular}} & \textbf{\begin{tabular}[c]{@{}l@{}}\# ADR not\\ Mentioned Tweets\end{tabular}} \\ \hline
Training         & 6725                                                               & 2                                                                   & 721                                                                        & 6004                                                                           \\ \hline
Validation           & 3535                                                               & 2                                                                   & 240                                                                        & 3295                                                                           \\ \hline
Testing          & 9961                                                               & 2                                                                   & 9190                                                                       & 771                                                                            \\ \hline
\end{tabular}
\end{table}

\begin{table}[!ht]
\centering
\caption{Task 2 Data Statistics}
\label{my-label}
\begin{tabular}{|l|l|l|l|l|l|}
\hline
\textbf{Data} & \textbf{\begin{tabular}[c]{@{}l@{}}Total \#\\ Tweets\end{tabular}} & \textbf{\begin{tabular}[c]{@{}l@{}}Total \#\\ Classes\end{tabular}} & \textbf{\begin{tabular}[c]{@{}l@{}}Personal\\ Medicine Intake\end{tabular}} & \textbf{\begin{tabular}[c]{@{}l@{}}Possible\\ Medicine Intake\end{tabular}} & \begin{tabular}[c]{@{}l@{}}Non\\ Intake\end{tabular} \\ \hline
Training         & 1065                                                               & 3                                                                   & 192                                                                         & 373                                                                         & 500                                                  \\ \hline
Validation           & 712                                                                & 3                                                                   & 125                                                                         & 230                                                                         & 357                                                  \\ \hline
Testing          & 7513                                                               & 3                                                                   & 1731                                                                        & 2697                                                                        & 3085                                                 \\ \hline
\end{tabular}
\end{table}

\subsection*{Results}
All experiments are trained using backpropogation through time (BPTT) \cite{ref14} on Graphics processing unit (GPU) enabled TensorFlow \cite{ref15} computational framework in conjunction with Keras framework in Ubuntu 14.04. We have submitted one run based on LSTM for task 1 and two runs composed of one run based on RNN and other one based on LSTM for task 2. The evaluation results is given by shared task committee are reported in Table 3 and 4.

\begin{table}[!ht]
\centering
\caption{Task 1 Results}
\label{my-label}
\begin{tabular}{|l|l|l|l|}
\hline
\textbf{Run} & \textbf{ADR Precision} & \textbf{ADR Recall} & \textbf{ADR F-score} \\ \hline
1            & 0.078                  & 0.17               & 0.107                \\ \hline

\end{tabular}
\end{table}

\begin{table}[!h]
\centering
\caption{Task 2 Results}
\label{my-label}
\begin{tabular}{|l|l|l|l|}
\hline
\textbf{Run} & \textbf{\begin{tabular}[c]{@{}l@{}}Micro-averaged precision\\ for classes 1 and 2\end{tabular}} & \textbf{\begin{tabular}[c]{@{}l@{}}Micro-averaged recall\\ for classes 1 and 2\end{tabular}} & \textbf{\begin{tabular}[c]{@{}l@{}}Micro-averaged F-score\\ for classes 1 and 2\end{tabular}} \\ \hline
1            & 0.414                                                                                           & 0.107                                                                                         & 0.171                                                                                         \\ \hline
2            & 0.843                                                                                           & 0.487                                                                                         & 0.617                                                                                         \\ \hline
\end{tabular}
\end{table}

\section*{Conclusion}
Social media mining is considerably an important source of information in many of health applications. This working note presents RNN and LSTM based embedding system for social media health text classification. Due to limited number of tweets, the performance of the proposed method is very less. However, the obtained results are considerable and open the way in future to apply for the social media health text classification. Moreover, the performance of the LSTM based embedding for task 2 is good in comparison to the task 1. This is primarily due to the fact that the target classes of task 1 data set imbalanced. Hence, the proposed method can be applied on large number of tweets corpus in order to attain the best performance.  

\makeatletter
\renewcommand{\@biblabel}[1]{\hfill #1.}
\makeatother

\bibliographystyle{unsrt}

\begin{thebibliography}{1}
\setlength\itemsep{-0.1em}

\bibitem{ref1} Lee, Kathy, Ankit Agrawal, and Alok Choudhary,
{Real-time disease surveillance using twitter data: Demonstration on flu and cancer}, In Proceedings of the 19th ACM SIGKDD International Conference on Knowledge Discovery and Data Mining, KDD ’13, pages 1474–1477, New York, NY, USA, 2013. ACM.

\bibitem{ref2} Lee, Kathy, Ankit Agrawal, and Alok Choudhary,
{Mining social media streams to improve public health allergy surveillance}, In 2015 IEEE/ACM International Conference on Advances in Social Networks Analysis and Mining (ASONAM), pages 815–822, Aug 2015.

\bibitem{ref3} Sarker, Abeed, and Graciela Gonzalez
{Portable automatic text classification for adverse drug reaction detection via multi-corpus training}, Journal of Biomedical Informatics, 53:196 – 207, 2015.

\bibitem{ref4} Jonnagaddala, J. I. T. E. N. D. R. A., TONI ROSE Jue, and H. J. Dai {Binary classification of Twitter posts for adverse drug reactions},  Proceedings of the Social Media Mining Shared Task Workshop at the Pacific Symposium on Biocomputing, Big Island, HI, USA. 2016.

\bibitem{ref5} Dai, Hong-Jie, Musa Touray, Jitendra Jonnagaddala, and Shabbir Syed-Abdul {Feature engineering for recognizing adverse drug reactions from twitter posts}, Information 7, no. 2 (2016): 27

\bibitem{ref6} Rastegar-Mojarad, M. A. J. I. D., Ravikumar Komandur Elayavilli, Yue Yu, and Hongfang Liu {Detecting signals in noisy data-can ensemble classifiers help identify adverse drug reaction in tweets} In Proceedings of the Social Media Mining Shared Task Workshop at the Pacific Symposium on Biocomputing. 2016.

\bibitem{ref7} Zhang, Zhifei, J. Y. Nie, and Xuyao Zhang {An ensemble method for binary classification of adverse drug reactions from social media},In Proceedings of the Social Media Mining Shared Task Workshop at the Pacific Symposium on Biocomputing. 2016.

\bibitem{ref8} Nguyen, Huy, and Minh-Le Nguyen {A Deep Neural Architecture for Sentence-level Sentiment Classification in Twitter Social Networking},arXiv preprint arXiv:1706.08032 (2017).

\bibitem{ref9} Lee, Kathy, Ashequl Qadir, Sadid A. Hasan, Vivek Datla, Aaditya Prakash, Joey Liu, and Oladimeji Farri {Adverse Drug Event Detection in Tweets with Semi-Supervised Convolutional Neural Networks}, In Proceedings of the 26th International Conference on World Wide Web, pp. 705-714. International World Wide Web Conferences Steering Committee, 2017.

\bibitem{ref10} Elman, Jeffrey L{Finding structure in time}, Cognitive science 14.2 (1990): 179-211.

\bibitem{ref11} Hochreiter, Sepp, and J\''urgen Schmidhuber {Long short-term memory}, Neural computation 9, no. 8 (1997): 1735-1780.

\bibitem{ref12} Gers, Felix A., Jürgen Schmidhuber, and Fred Cummins {Learning to forget: Continual prediction with LSTM},(1999): 850-855.

\bibitem{ref13} Gers, Felix A., Nicol N. Schraudolph, and Jürgen Schmidhuber {Learning precise timing with LSTM recurrent networks}, Journal of machine learning research 3.Aug (2002): 115-143.

\bibitem{ref14} Werbos, Paul J {Backpropagation through time: what it does and how to do it},  Proceedings of the IEEE 78.10 (1990): 1550-1560.

\bibitem{ref15} Abadi, M., Barham, P., Chen, J., Chen, Z., Davis, A., Dean, J., Devin, M., Ghemawat, S., Irving, G., Isard, M. and Kudlur, M. {TensorFlow: A System for Large-Scale Machine Learning}, In OSDI, vol. 16, pp. 265-283. 2016.

\bibitem{ref15} Barathi Ganesh HB, Anand Kumar M and Soman KP. "Distributional Semantic Representation in Health Care Text Classification." (2016).

\end{thebibliography}

\end{document}